\documentclass[
hf,
]{ceurart}

\usepackage{booktabs} %
\usepackage{float}
\restylefloat{table}
\usepackage{enumitem}
\usepackage{multirow}
\usepackage{colortbl}

\usepackage[most]{tcolorbox}

\newcommand{\code}[1]{{\rmfamily#1}}

\definecolor{codegreen}{rgb}{0,0.6,0}
\definecolor{codegray}{rgb}{0.5,0.5,0.5}

\definecolor{backcolour}{RGB}{245,248,250}
\definecolor{emph}{RGB}{166,88,53}
\definecolor{nightblue}{RGB}{9,49,105}
\definecolor{keywords}{RGB}{207,33,46}
\definecolor{lightpurple}{RGB}{130,81,223}

\lstdefinestyle{mystyle}{
    backgroundcolor=\color{backcolour},   
    commentstyle=\color{codegreen},
    keywordstyle=\color{keywords},
    stringstyle=\color{nightblue},
    basicstyle=\fontsize{7}{8}\ttfamily,
    breakatwhitespace=true,         
    breaklines=true,                 
    captionpos=b,                    
    keepspaces=true,                 
    numberstyle=\tiny\color{codegray},
    numbersep=2pt,                  
    showspaces=false,                
    showstringspaces=false,
    showtabs=false,                  
    tabsize=2,
    moredelim=**[is][\color{red}]{@}{@},
    moredelim=**[is][\ttfamily\color{black}]{+}{+},
    emph={dsp,Example,sample,annotate,knn,crossval,generate,retrieve,retrieve\_ensemble,majority,fused_retrieval,Template, Transformation,rank,branch},
    emphstyle={\color{lightpurple}},
    linewidth=0.98\columnwidth,
    frame=tb,    
    xrightmargin=0pt,
    xleftmargin=0.23cm,
    numbers=left,
    aboveskip=0.4cm,
    belowskip=0.4cm,
}

\begin{document}

\conference{Workshop on Causal Neuro-symbolic Artificial Intelligence, June 01--5, 2025, Portoroz, Slovenia}

\title{Prompting or Fine-tuning? Exploring Large Language Models for Causal Graph Validation}

\author[1]{Yuni Susanti}[%
email=susanti.yuni@fujitsu.com,
]
\address[1]{Artificial Intelligence Dept., Fujitsu Ltd., Japan}

\author[2]{Nina Holsmoelle}[%
email=n.holsmoelle@gmail.com
]
\fnmark[1]
\address[2]{Osaka University, Japan}

\fntext[1]{Work performed while at Fujitsu Limited (internship).}

\begin{abstract}
This study explores the capability of Large Language Models (LLMs) to evaluate causality in causal graphs generated by conventional statistical causal discovery methods—a task traditionally reliant on manual assessment by human subject matter experts. To bridge this gap in causality assessment, LLMs are employed to evaluate the causal relationships by determining whether a causal connection between variable pairs can be inferred from textual context. Our study compares two approaches: (1) prompting-based method for zero-shot and few-shot causal inference (\textit{unsupervised}) and, (2) fine-tuning language models for the causal relation prediction task (\textit{supervised}). While prompt-based LLMs have demonstrated versatility across various NLP tasks, our experiments on biomedical and general-domain datasets show that fine-tuned models consistently outperform them, achieving up to a 20.5-point improvement in F1 score—even when using smaller-parameter language models. These findings provide valuable insights into the strengths and limitations of both approaches for causal graph evaluation. 

\end{abstract}

\maketitle
\vspace{-0.3cm}
\section{Introduction}
\label{sec:intro}
\begin{wrapfigure}[13]{l}{4cm}
\centering
\vspace*{-0.2\baselineskip}%
  \includegraphics[width=3cm]{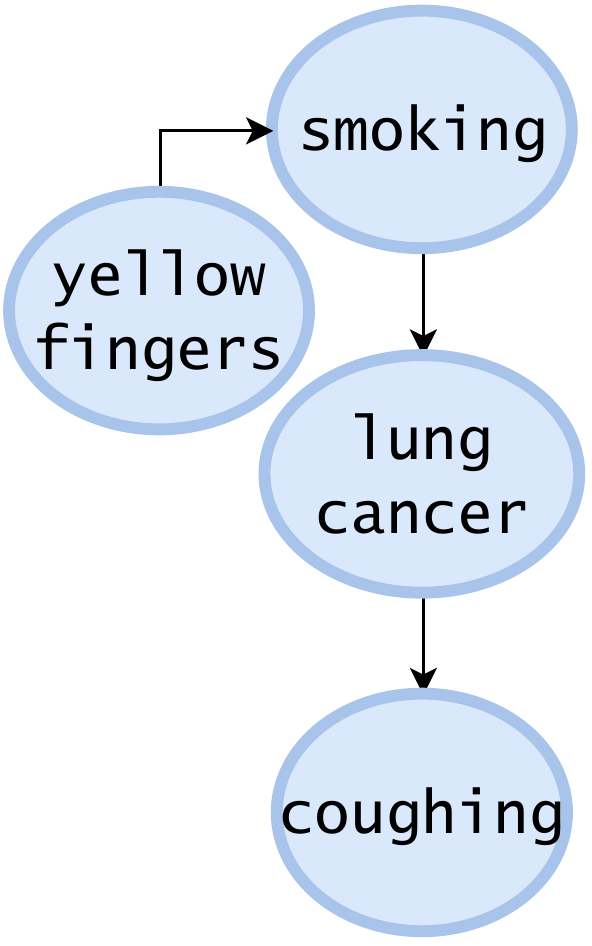}
  \vspace{-0.4cm}
  \caption{Illustration of a causal graph.}
  \label{fig:causalgraph}
\end{wrapfigure}

\vspace{-0.3cm}
Uncovering underlying causal relationships is a fundamental task across various scientific disciplines, as these relationships form the basis for understanding and decision-making. Statistical causal discovery methods~\cite{Spirtes2000,Lingam10.5555/1248547.1248619} estimate causal structures from observational data, generating causal graphs that visualize these relationships, as illustrated in Figure~\ref{fig:causalgraph}.

Despite significant advancements in causal discovery, a major challenge remains: verifying the accuracy of causal graphs produced by these predominantly \textit{unsupervised} methods. Typically, this verification relies on domain experts manually validating the graphs, often through controlled experiments. However, such experiments can be prohibitively expensive or, in some fields, entirely unfeasible due to ethical constraints. This highlights the pressing need for alternative, scalable methods to verify causal graphs.

Another approach to verifying causal graphs is using external knowledge from text sources. Causal information is widely distributed across diverse sources, making it an invaluable resource for assisting human experts in validating the accuracy of causal graphs. However, as the number of variables in a causal graph grows and the volume of textual information expands rapidly, manual verification becomes increasingly impractical. Natural Language Processing (NLP) technologies, including Large Language Models (LLMs) like BERT~\cite{BERTDBLP:journals/corr/abs-1810-04805} and ChatGPT, offer a promising solution. These models infer causal relationships between node pairs by leveraging their pre-trained knowledge to analyze the relevant textual context. 

In this work, we examine the feasibility of applying NLP technologies to automate causal graph verification. Through quantitative evaluation on causal text datasets, we investigate the performance of two distinct types of NLP models: (1) pre-trained language models fine-tuned for the task of causal relation classification (\textit{supervised}), and (2) prompt-based LLMs (\textit{unsupervised}). To sum up the results, prompt-based LLMs do not necessarily perform better than supervised models on this task, despite their promising performance on diverse clinical NLP tasks~\cite{agrawal-etal-2022-large}. We conduct a detailed analysis to explore the potential factors contributing to this performance gap. Our findings offer valuable insights into the strengths and limitations of these approaches for scalable, automated causal graph validation.

\vspace{-0.3cm}
\section{Related Work}
\label{relwork}
\vspace{-0.3cm}
The research on causal relation extraction/classification from text sources has been done mostly in supervised setting, especially in biomedical-chemistry domains~\cite{Khoo2.10.3115/1075218.1075261,Mihăilă2014,Gu10.1093/database/baw042,medicause,khetan-etal-2022-mimicause,SusantiCEG}, and open-domain~\cite{khoo1998,CHANG2006662,blanco-etal-2008-causal,balashankar-etal-2019-identifying}. The pre-training and fine-tuning paradigm in NLP led to state-of-the-art performance in many downstream tasks; likewise, most of the related works listed above fine-tune the pre-trained language models such as BERT~\cite{devlin-etal-2019-bert}, or propose some sort of enhancement for BERT such as the work by~\cite{Su2022,SusantiCEG}. Their results on relation extraction on biomedical datasets has been encouraging, motivating us to choose BERT as the model for our fine-tuning experiments. On the other hand, recent works~\cite{kic2023causal,tu2023causaldiscovery,willig2022foundation,zhang2023understanding,gao-etal-2023-chatgpt,SusantiKGSP} show that  Large Language Models (LLMs) effectively provide background knowledge for causal discovery, and their findings suggest that LLM-based prompting methods achieved superior performance than non-LLM approaches. For instance,~\cite{agrawal-etal-2022-large} shows that LLMs perform well at zero and few-shot information extraction from clinical text, despite not being trained specifically for the clinical domain. Similarly, other works~\cite{wei2023zeroshot,jeblick2022chatgpt} suggest that LLMs (i.e., InstructGPT~\cite{ouyang2022training}, ChatGPT, GPT-3.5, etc.), perform well in various downstream tasks even without tuning the parameters, but only with few examples as instructions/prompts. This inspires us to evaluate such instruction, or prompt-based LLMs, on our causal relation classification task. In this work, we compare the prompt-based LLMs against the \textit{more traditional} supervised model where it is trained/fine-tuned using the training data for causal relation classification task.

\vspace{-0.3cm}
\section{Approach}
\label{sec:approach}
\vspace{-0.3cm}
Given a pair of entities $e_{1}$ and $e_{2}$ (i.e., node pairs in causal graph such as \textit{smoking} and \textit{lung cancer}), the LLM is tasked with determining whether a causal relationship exists between them. This formulation frames the problem as a \textit{classification} task, where the relation is categorized as either \textit{causal} or \textit{non-causal}. We explore both prompt-based and fine-tuned LLMs, as below:

\vspace{-0.3cm}
\subsection{Prompt-based LLMs}
\label{sec:prompt} 
\vspace{-0.3cm}
In prompt-based learning, a pre-trained language model is adapted
to a specific task via priming on natural language prompts—pieces of text that are combined with an input and then fed to the language model to produce an output for that task~\cite{agrawal-etal-2022-large}. Prompt-based learning requires the specification of a prompt template to be applied to the input, thus we designed two settings for the prompt-based LLMs experiments: \textbf{Single-Prompt} and \textbf{Few-Shot Prompt} settings.

\vspace{-0.3cm}
\paragraph{\textbf{Single-Prompt.}}
\label{sec:prompt1} 
In the \code{Single-Prompt} setting, we instruct the prompt to directly ask the LLMs to answer a question about causality between a pair of entities, without providing any example of the training data in the prompt (i.e., \textit{zero-shot} approach). For the pair $e_{1}$ and $e_{2}$ with textual context $S$, we hand-crafted the following three prompt variations.
\begin{tcolorbox}
    [
    title=\textbf{A: two-choices, no-context}, 
    colback=gray!5, 
    colframe=black, 
    fonttitle=\bfseries,
    boxsep=3pt, 
    left=3pt, 
    right=3pt, 
    top=3pt, 
    bottom=3pt,
    sharp corners ]
There is a causal relationship between $e_{1}$ and $e_{2}$. Answer with `True' or `False'  
\end{tcolorbox}

\begin{tcolorbox} [
    title=\textbf{B: two-choices, with-context}, 
    colback=gray!5, 
    colframe=black, 
    fonttitle=\bfseries,
    boxsep=3pt, 
    left=3pt, 
    right=3pt, 
    top=3pt, 
    bottom=3pt,
    sharp corners ]
    Given the following context, classify the relationship between $e_{1}$ and $e_{2}$ as causal or non-causal. Answer with `causal' or `non-causal'. Context: $S$
\end{tcolorbox}

\begin{tcolorbox} [
    title=\textbf{C: three-choices, with-context}, 
    colback=gray!5, 
    colframe=black, 
    fonttitle=\bfseries,
    boxsep=3pt, 
    left=3pt, 
    right=3pt, 
    top=3pt, 
    bottom=3pt,
    sharp corners ]
    Given the context below, is there a causal relationship between  $e_{1}$ and $e_{2}$. In case only a correlation, but no strict causation between $e_{1}$ and $e_{2}$, answer with `False.' In case of uncertainty, answer with `Maybe.' In a case where there is clearly a causal relationship, and not just a correlation between $e_{1}$ and $e_{2}$, answer with `True.'  Context: $S$
\end{tcolorbox}

The LLMs are strictly constrained to respond with two choices, as in variations (A) and (B), ensuring a fair comparison with the fine-tuned model. However, in variation (C), we allow the LLMs to return a "Maybe" option when they indicate insufficient evidence to determine causality. 
Additionally, we varied the prompt by either \textit{including} or \textit{omitting} the textual context sentence $S$, referred to as \textbf{with-context} and \textbf{no-context}, respectively.

\vspace{-0.3cm}
\paragraph{\textbf{Few-Shot Prompt.}}
\label{sec:prompt2} In the \texttt{Few-Shot} prompt setting, we structured the prompt to include \( n \) examples of training data, allowing the LLMs to process and learn from these examples before making predictions. This method also often referred as in-context learning~\cite{GPT3NEURIPS2020_1457c0d6}. Each training example contains:  (a) the entity pair \( e_1 \) and \( e_2 \), (b) the relation label (\textit{causal} or \textit{non-causal}), and (c) the sentence \( S \) as context. As shown below, the LLM is tasked to classify the test data:

\begin{tcolorbox}[
    colback=gray!5, 
    colframe=black, 
    fonttitle=\bfseries,
    boxsep=3pt, 
    left=3pt, 
    right=3pt, 
    top=3pt, 
    bottom=3pt,
    sharp corners ]
    Given the context sentence, classify the relationship between the entities marked with $e_{1}$ and $e_{2}$ as \textit{causal} or \textit{non-causal} \\
    \textcolor{blue}{\textbf{Context Sentence}: Expression of <e1> osteopontin </e1> contributes to the progression of <e2> prostate cancer </e2>.
    \textbf{Result}: e1: \textit{osteopontin}, relation: \textit{causal}, e2: \textit{prostate cancer}'} \\
    \textcolor{red}{\textbf{Context Sentence}: Increased expression of <e1> cyclin B1 </e1> sensitizes <e1> prostate cancer </e1> cells to apoptosis induced by chemotherapy. \textbf{Result}: }
    
    \end{tcolorbox}

Here, one training example (\( n = 1 \)) is embedded in the prompt, highlighted in \textcolor{blue}{blue}, while the test example is marked in \textcolor{red}{red}. 
The expected output is shown below:
\begin{tcolorbox}[boxsep=2pt,left=2pt,right=2pt,top=2pt,bottom=2pt]
\textcolor{red}{e1: \textit{cyclin B1}, relation: \textit{causal}, e2: \textit{prostate cancer}}
\end{tcolorbox}

We conducted the \code{Few-Shot Prompt} experiment by varying the number of the training data \textit{n} to be included in the prompt. 

\vspace{-0.3cm}
\subsection{Fine-tuned LLMs}
\label{sec:finetuning}
\vspace{-0.3cm}
The pre-training of LLMs usually utilizes a great quantity of unlabeled data, and the fine-tuning involves training these pre-trained LLMs on a smaller dataset labeled with examples relevant to the target task. By exposing the model to these new labeled examples, the model adjusts its parameters and internal representations suited for the target task. In this work, we experimented with two models: \textbf{(1) BERT}~\cite{BERTDBLP:journals/corr/abs-1810-04805} to represent \textbf{\textit{small} language model} (under 1b parameters) and \textbf{(2) GPT} to represent \textbf{\textit{larger}-parameter models}.

\vspace{-0.3cm}
\paragraph{\textbf{Fine-tuning BERT model}}
\label{sec:finetune}

\begin{wrapfigure}[18]{l}{6.8cm}
\centering
\vspace*{-0.2\baselineskip}%
  \includegraphics[width=6.5cm]{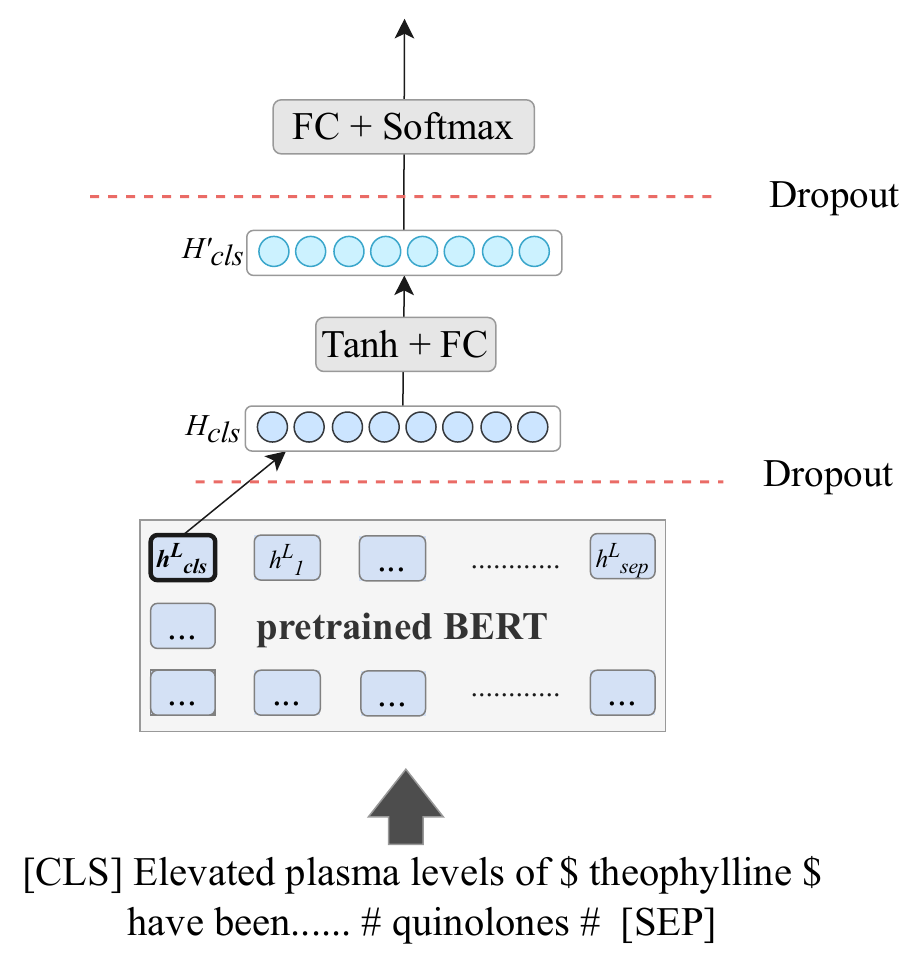}
  \caption{Fine-tuning BERT.}
  \label{fig:modelarc}
\end{wrapfigure}

BERT is a language model trained using a multi-layer bidirectional transformer encoder and has been demonstrated to enhance various NLP tasks~\cite{BERTDBLP:journals/corr/abs-1810-04805}. Figure~\ref{fig:modelarc} illustrates our model architecture for fine-tuning the pre-trained BERT model on the causal relation classification task. To ensure a fair comparison with prompt-based LLMs, we adopt a simple fine-tuning approach. Given an input sequence $S$ , we extract its vector representation from the BERT model and utilize the last hidden state of the [CLS] token as the input representation for fine-tuning, following the original paper ~\cite{BERTDBLP:journals/corr/abs-1810-04805}.
We further apply a $Tanh$ activation function and a fully-connected layer (FC) to this representation to obtain the final sequence representation $H'_{cls}$:
\vspace{-0.1cm}
\begin{equation}
    H'_{cls} = W_{0}(tanh(H_{cls}))+b_{0} 
\end{equation}

Dropout is applied in the model architecture as a regularization method, as indicated in Figure~\ref{fig:modelarc}. We used the \textit{binary cross entropy} as the loss function during the training.

\vspace{-0.3cm}
\paragraph{\textbf{Fine-tuning GPT model}}
\label{sec:finetuning-gpt} 
Fine-tuning the GPT model includes formatting each training example into \textit{\textbf{prompt-completion}} pair, where the input example serves as the \textbf{prompt}, and the corresponding output serves as the \textbf{completion}. The format of these pairs varies depending on the task. While our task is fundamentally a relation \textit{classification} task, it can also be framed as a relation \textit{extraction} task between pairs of entities. We followed GPT fine-tuning instruction
and formatted the examples into both task formats:.
\begin{tcolorbox}
    [
    title=\textbf{A: Fine-tuning GPT, \textbf{classification} format}, 
    colback=gray!5, 
    colframe=black, 
    fonttitle=\bfseries,
    boxsep=3pt, 
    left=3pt, 
    right=3pt, 
    top=3pt, 
    bottom=3pt,
    sharp corners ]
    {\textbf{Prompt}: \textcolor{blue}{The results provide evidence for altered plasticity of synaptic morphology in memory mutants <e1>dnc</e1> and <e2>rut</e2> and suggest a role...}\\
    \textbf{Completion}:  \textcolor{red}{non-causal END}}
\end{tcolorbox}

\begin{tcolorbox}
    [
    title=\textbf{B: Fine-tuning GPT, \textbf{relation extraction} format}, 
    colback=gray!5, 
    colframe=black, 
    fonttitle=\bfseries,
    boxsep=3pt, 
    left=3pt, 
    right=3pt, 
    top=3pt, 
    bottom=3pt,
    sharp corners ]
    {\textbf{Prompt}: \textcolor{blue}{The results provide evidence for altered plasticity of synaptic morphology in memory mutants <e1>dnc</e1> and <e2>rut</e2> and suggest a role...}\\
    \textbf{Completion}:  \textcolor{red}{dnc rut non-causal END}}
\end{tcolorbox}

\vspace{-0.4cm}
\section{Evaluation}
\vspace{-0.3cm}
\subsection{Datasets and Experiment Settings}
\label{sect:data}
\vspace{-0.3cm}
We evaluate our approach in (A) \textbf{biomedical domain}, focusing on 3 types of causality: gene-gene (\code{GENEC}~\cite{SusantiKGSP}),  drug-side effect (\code{DDI}~\cite{ddiHERREROZAZO2013914}), and gene-disease (\code{COMAGC}~\cite{Lee2013com}), and (B) \textbf{open-domain} causality dataset \code{SEMEVAL}~\cite{hendrickx-etal-2010-semeval}. We used GPT model using OpenAI API with \texttt{gpt-3.5-turbo} and \texttt{text-davinci-003} engines. For experiments with BERT model, we applied \code{BioBERT}~\cite{BioBERT10.1093/bioinformatics/btz682}, \code{PubMedBERT}~\cite{PubmedBERT10.1145/3458754} for biomedical dataset and \code{BERT} (\textit{large, uncased}) for open-domain dataset. Code, dataset and hyper-parameters settings are provided in our Github. 

\vspace{-0.3cm}
\subsection{Results and Discussion}
\label{sec:resultdis}
\vspace{-0.3cm}
Table~\ref{tab:semresult} summarizes the evaluation results for the biomedical and open-domain datasets. We report the Precision (P), Recall (R), and F1 scores. We apply 5-fold cross-validation and the scores are averaged. We report the standard deviation values of the F1 scores over the 5-folds as shown in parenthesis in Table~\ref{tab:semresult}. 

\begin{table*}
    \centering
    \footnotesize
    \tabcolsep=0.10cm
    \begin{tabular}{lccccccccccccc}
    \toprule
     & & \multicolumn{3}{c}{(Biomed) \code{COMAGC}} & \multicolumn{3}{c}{(Biomed) \code{DDI}} & \multicolumn{3}{c}{(Biomed) \code{GENE}} & \multicolumn{3}{c}{(News) \code{SEMEVAL}}\\
    \textbf{Prompt-based} & $type$ & P & R & F1 & P & R & F1 & P & R & F1 & P & R & F1\\
    \midrule
    \code{Single-Prompt} & A & 28.2 & 61.0 & 38.1 (.06) & 52.2 & 25.7 & 34.3 (.02)  
    & 23.6 & 26.6 & 24.2 (.07) & 64.6 & 66.0 & 65.3 (.06) \\
    \code{Single-Prompt} & B & 28.2 & 94.2 & 43.2 (.05)& 65.1 & 69.0 & 66.7 (.04) 
    & 34.3 & 59.6 & \textbf{42.3} (0.1) & 77.7 & 84.7 & \textbf{80.8} (.04)\\
    \code{Single-Prompt} & C & 48.8 & 100 & \textbf{64.2} (.14) & 52.9 & 93.2 & \textbf{67.4} (.02)
    & 27.4 & 71.9 & 39.5 (.05) & 57.4 & 82.8 & 67.7 (.03)\\
    \midrule
    & $n$ &  &  & \\
    \midrule
    \code{Few-Shot Prompt} & 5 & 37.2 & 83.5 & 51.0 (.03)& 100 & 37.6 & \textbf{53.1} (.15)
    & 22.1 & 25.7 & 22.7 (.28) & 100 & 46.0 & 62.7 (.06)\\
    \code{Few-Shot Prompt} & 15 & 52.8 & 41.4 & 46.1 (.08) & 51.4 & 27.0 & 35.1 (.05)
    & 26.0 & 29.1 & 26.2 (.18) & 100 & 47.9 & \textbf{64.6} (.04)\\
    \code{Few-Shot Prompt} & 20 & 50.2 & 70.4 & \textbf{57.0} (.08)& * & * & *
    & 31.7 & 39.5 & \textbf{34.3} (.08) & 58.9 & 57.7 & 58.2 (.02)\\
    \midrule
    \textbf{Fine-tuning} &  &  &  & \\
    \midrule 
    \code{BioBERT} &  & 77.9 & 84.4 & 80.8 (.01)& 97.0 & 76.2 & 85.2 (.03) 
    & 46.1 & 65.2 & 53.5 (.07*) & * & * & * \\
    \code{PubmedBERT} &  & 80.7 & 87.4 & \textbf{83.9} (.03)& 93.2 & 83.3 & \textbf{87.9} (.01) 
    & 50.6 & 62.1 & \textbf{55.1} (.03) & * & * & * \\
    \code{BERT-large} &  & * & * & * & * & * & * & * & * & * & 93.0 & 93.0 & 93.0 (.01) \\
     \midrule
    \code{GPT} (\textit{classification}) & & 80.5 & 70.1 & 74.1 (.06) & 99.4 & 78.1 & 87.4 (.03) 
    & 58.6 & 23.1 & 31.4 (.08) & 99.9 & 94.8 & \textbf{96.8} (.03)\\
    \code{GPT} (\textit{extraction}) & & 75.6 & 58.1 & 65.5 (.07) & 100 & 62.9 & 77.1 (.02)
     & 52.4 & 21.2 & 30.1 (.06) & 100 & 91.9 & 95.7 (.03)\\
    \bottomrule
    \end{tabular}
    \caption{Experiment results. Values in \textbf{bold} indicates the best F1 score for each method and dataset. $type$ refers to Single-Prompt variations as explained in~\ref{sec:prompt1}, $n$ refers to the number of training data included in the prompt for Few-Shot setting.}
    \label{tab:semresult}
\end{table*}

In summary, the results indicate that fine-tuned LLM models significantly outperform prompt-based LLMs, achieving a 12.8–20.5 point improvement in F1 score across all datasets. Notably, even with smaller language models like BERT, fine-tuning leads to a substantial performance boost, with F1 scores improving by up to 20.5 points (67.4 vs. 87.9, \code{Single-Prompt C} vs. \code{PubMedBERT} on \code{DDI} dataset). This contrasted with the previous studies~\cite{wei2023zeroshot,jeblick2022chatgpt} where LLMs perform relatively well, if not better than the fine-tuned models in various tasks including in clinical NLP tasks~\cite{agrawal-etal-2022-large}. 
One possible reason that the prompt-based model does not perform as well as the fine-tuned model is that causality is rarely written explicitly with causal cues like \textit{“cause,” “causing,” or “caused”}. Instead, it is often described more implicitly or ambiguously, using keywords such as \textit{“contribute”} or \textit{“play a role”}. Additionally, by fine-tuning the model with training samples, we expose the model to various ways in which causal relationships can be expressed in text. This suggests that identifying causality patterns from training samples is a crucial step in accurately recognizing causal relations.
Nevertheless, in the \code{Few-Shot Prompt} experiments, where $n$ training samples are included in the prompt, the performance does not always improve compared to models without training samples. For example, the scores are 67.4 vs. 53.1 with \code{Single-Prompt C} versus \code{Few-Shot Prompt} with $n$=5 on the \code{DDI} dataset. This is illustrated in Table~\ref{tab:semresult}, where, surprisingly, the highest F1 score for the prompt-based methods is achieved with \code{Single-Prompt} B and C, both of which do not include any training samples. We hypothesize that the limited size of the training samples may be a factor, and increasing the amount of training data could potentially improve results. However, due to the token limitations of the OpenAI API, we were unable to experiment with larger values of $n$.

Next, we investigated the effect of including the context sentence in the prompt. To do this, we created prompt variations of the Single-Prompt model by \textit{including} and \textit{not including} the context sentence $S$ in the prompt (\textbf{with-context} and \textbf{no-context}).
The results suggest that, for prompt-based methods, including the context sentence in the prompt can be effective. As shown in Table~\ref{tab:semresult}, the \code{Single-Prompt} types B and C (\textbf{with-context}) consistently outperform type A (\textbf{no-context}). By incorporating context, the model gains additional knowledge to better predict the relationship between the entity pair, rather than relying solely on the information acquired during pre-training. In addition, we observed generally higher scores on the open-domain dataset (96.8 with fine-tuned \code{GPT} on \code{SEMEVAL}) compared to the biomedical datasets (87.9 with \code{PubMedBERT} on \code{DDI}). This is expected, as LLMs are predominantly pre-trained on open-domain texts, such as books, articles, and online content. Another contributing factor could be the complexity of biomedical texts, which often include more domain-specific or technical terms compared to open-domain datasets.

\vspace{-0.3cm}
\section{Conclusion}
\label{sec:conclusion}
\vspace{-0.3cm}
We present a study exploring the feasibility of applying NLP technologies for causal graph verification. Specifically, we compare prompt-based and fine-tuned LLMs in predicting causality between pairs of entities. Experiments on biomedical and open-domain datasets suggest that fine-tuned models outperform prompt-based LLMs, even with smaller-parameter models like BERT. However, fine-tuned models require sufficient expert-annotated data for training, which can be a significant bottleneck. Constructing training data through expert annotation is often challenging and costly. In this regard, LLMs hold promise as a breakthrough for causal inference research. Due to data limitations, our current evaluation was restricted to binary pairwise causality assessment. Expanding this to analyze causal graphs with multiple interconnected variables is a key focus of our future work.

\bibliography{kr-sample}

\end{document}